\documentclass[final,3p,times]{elsarticle} 

\usepackage{amssymb}
\usepackage{subcaption}
\usepackage{caption}
\usepackage{subcaption}
\usepackage{graphicx}
\usepackage{multirow}
\usepackage{lscape}
\usepackage{color, colortbl}
\usepackage{xcolor}         
\usepackage{wasysym}
\usepackage{array}

\newcolumntype{C}[1]{>{\centering\arraybackslash}p{#1}}
\definecolor{CHANNEL}{HTML}{D5C2FF}
\definecolor{KERNEL}{HTML}{FFFCBD}
\definecolor{ORANGE}{HTML}{FFC3DE}
\definecolor{HYBRID}{HTML}{C3DFCA}

\journal{arXiv}

\begin{document}

\begin{frontmatter}



\title{Model Compression Methods for YOLOv5: A Review}


\author[inst1]{Mohammad Jani}
\ead{mhmdjani@uvic.ca}

\author[inst2]{Jamil Fayyad}
\ead{jfayyad@mail.ubc.ca}

\author[inst1]{Younes Al-Younes}
\ead{yalyounes@uvic.ca}

\author[inst1, corref]{Homayoun Najjaran}
\ead{najjaran@uvic.ca}

\affiliation[inst1]{organization={University of Victoria},
            addressline={800 Finnerty Road}, 
            city={Victoria},
            postcode={V8P 5C2}, 
            state={BC},
            country={Canada}}

\affiliation[inst2]{organization={The University of British Columbia},
            addressline={3333 University Way}, 
            city={Kelowna},
            postcode={V1V 1V7}, 
            state={BC},
            country={Canada}}

\cortext[corref]{Corresponding Author}
\begin{abstract}

Over the past few years, extensive research has been devoted to enhancing YOLO object detectors. Since its introduction, eight major versions of YOLO have been introduced with the purpose of improving its accuracy and efficiency. While the evident merits of YOLO have yielded to its extensive use in many areas, deploying it on resource-limited devices poses challenges. To address this issue, various neural network compression methods have been developed, which fall under three main categories, namely network pruning, quantization, and knowledge distillation.
The fruitful outcomes of utilizing model compression methods, such as lowering memory usage and inference time, make them favorable, if not necessary, for deploying large neural networks on hardware-constrained edge devices. 
In this review paper, our focus is on pruning and quantization due to their comparative modularity. We categorize them and analyze the practical results of applying those methods to YOLOv5. 
By doing so, we identify gaps in adapting pruning and quantization for compressing YOLOv5, and provide future directions in this area for further exploration. Among several versions of YOLO, we specifically choose YOLOv5 for its excellent trade-off between recency and popularity in literature.
This is the first specific review paper that surveys pruning and quantization methods from an implementation point of view on YOLOv5. Our study is also extendable to newer versions of YOLO as implementing them on resource-limited devices poses the same challenges that persist even today. This paper targets those interested in the practical deployment of model compression methods on YOLOv5, and in exploring different compression techniques that can be used for subsequent versions of YOLO.

\end{abstract}



\end{frontmatter}

\section{Introduction}
\label{sec:sample1}
As a fundamental problem, object detection has been an active area of research for many years. The main goal of object detection involves identifying and localizing objects of interest from different categories within some given image. Object detection is the basis of many other advanced computer vision tasks \cite{salari2022object} ranging from semantic segmentation \cite{hao2020brief} to object tracking \cite{ciaparrone2020deep} to activity recognition \cite{gonzalez2015features}. 
In recent years, deep learning-based approaches such as Convolutional Neural Networks (CNNs) have achieved state-of-the-art performance in object detection tasks. As a result of the advancements in computational power and cutting-edge algorithms, object detection has become more accurate, enabling a wide range of real-world applications.
Compared to classical object detection methods, using CNNs alleviates the problem of feature extraction, classification, and localization in object detection \cite{YOLO2016, girshick2014rich, girshick2015fast, ren2015faster, krizhevsky2017imagenet, zhao2019objSurvey}.

Typically, Object detection can be done through two methods, namely, single-stage and two-stage detection. While in the former, the algorithm directly predicts the bounding boxes and class probabilities for objects, in the latter, the algorithm first generates a set of region proposals and then classifies those proposals as objects or backgrounds \cite{jiao2019objSurvey}.
Unlike Faster R-CNN \cite{girshick2014rich} and R-FCN \cite{Rfcn2016r} as two-stage object detection methods, single-stage ones, like YOLO \cite{YOLO2016}, SSD \cite{ssd2016}, EfficientDet \cite{tan2020efficientdet}, and RetinaNet \cite{lin2017RetinaNet}, typically use one Fully Convolutional Neural Network (FCN) to detect objects' classes and spatial locations without intermediate steps. 

Among different single-stage object detection methods, YOLO has gained a lot of attention since it was published in 2016. The primary idea behind YOLO is to divide an input image into a grid of cells and predict bounding boxes and class probabilities for each cell. YOLO treats object detection as a  regression problem. Also, since it uses a single neural network for object detection and classification, it can be optimized jointly for both tasks, which ultimately enhances the whole detection performance. YOLOv1 was equipped with a simple structure containing 24 convolutional layers and two fully connected layers at the end to deliver probabilities and coordinates \cite{YOLO2016}.
Since its introduction, YOLO has undergone several improvements and variations. In 2017, YOLOv2 (aka YOLO9000) was published with improvements in the performance led by using multi-scale training, anchor boxes, batch normalization, Darknet-19 architecture, and a modified loss function \cite{redmon2017yolo9000}. Following that, Redmon and Farhadi introduced YOLOv3 that employs a feature pyramid network, convolutional layers with anchor boxes, Spatial Pyramid Pooling (SPP) block, Darknet-53 architecture, and an improved loss function \cite{redmon2018yolov3}. Unlike previous versions, YOLOv4 was introduced by different authors. A. Bochkovskiy et al. enhanced YOLO's performance by utilizing CSPDarknet53 architecture, Bag-of-Freebies, Bag-of-Specials, mish activation function, Weighted-Residual-Connections (WRC), Spatial Pyramid Pooling (SPP), and Path Aggregation Network (PAN) \cite{bochkovskiy2020yolov4}.

\begin{figure*}[t]
    \centering
    \includegraphics[width=\linewidth]{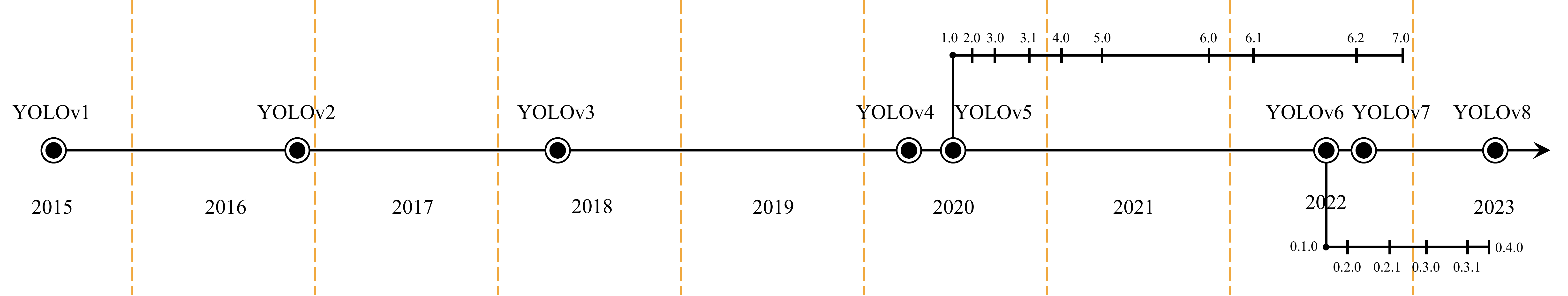}
    \caption{\textbf{YOLO release timeline.} YOLOv5 and YOLOv6 have ten and six released variants, respectively.}
    \label{fig:timeline}
\end{figure*}

\begin{figure}[t]
    \centering
    \includegraphics[width=\linewidth]{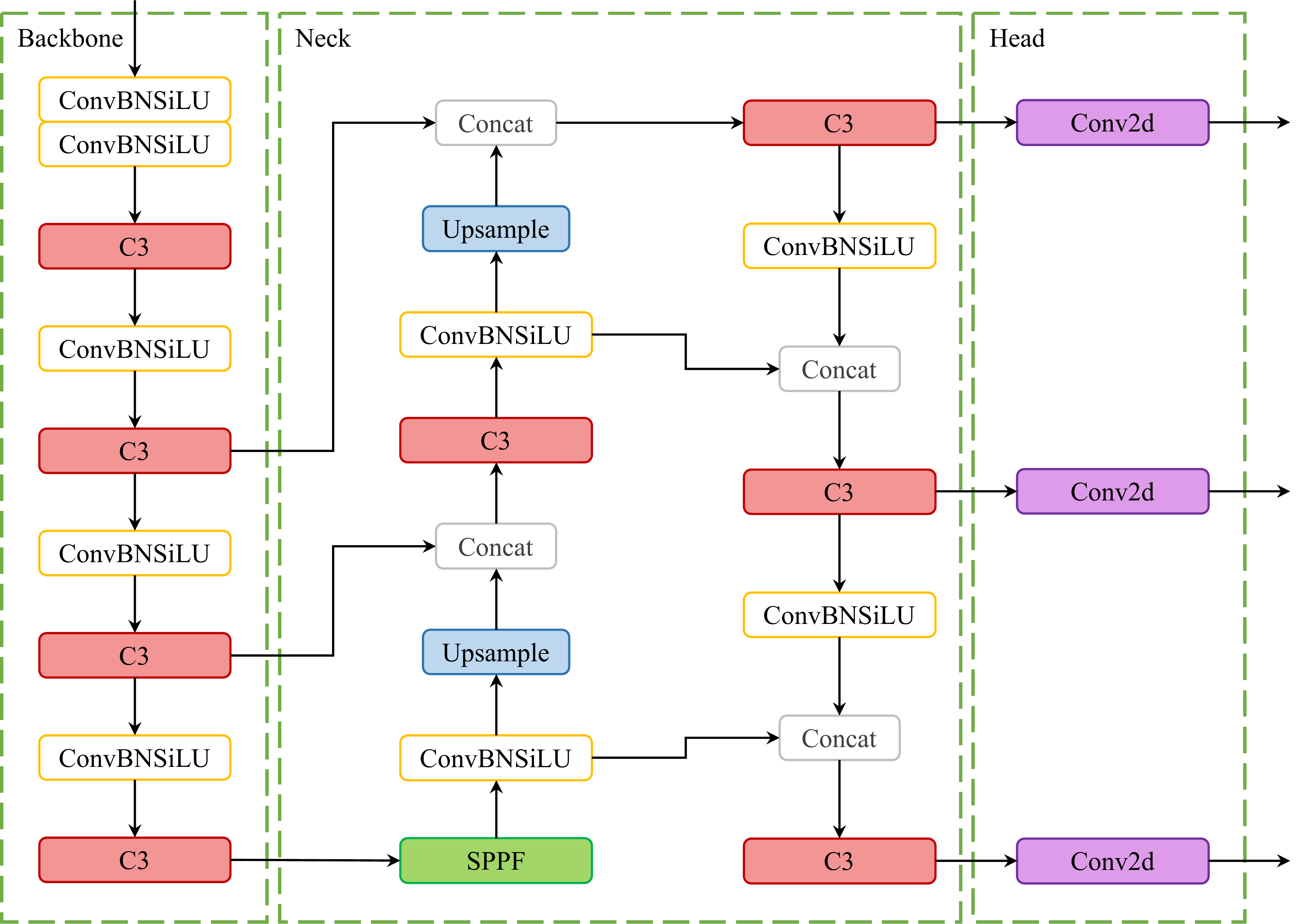}
    \caption{\textbf{YOLOv5l architecture.} SPPF represents a computation-efficient version of the Spatial Pyramid Pooling, which was originally implemented in YOLOv3; C3 uses the new CSP-combined module whose details are illustrated in Figure \ref{fig:C3_strc}.}
    \label{fig:YOLOv5l_whole}
\end{figure}

\begin{figure*}
    \centering
    \begin{subfigure}[b]{0.32\linewidth}
        \includegraphics[width=\linewidth]{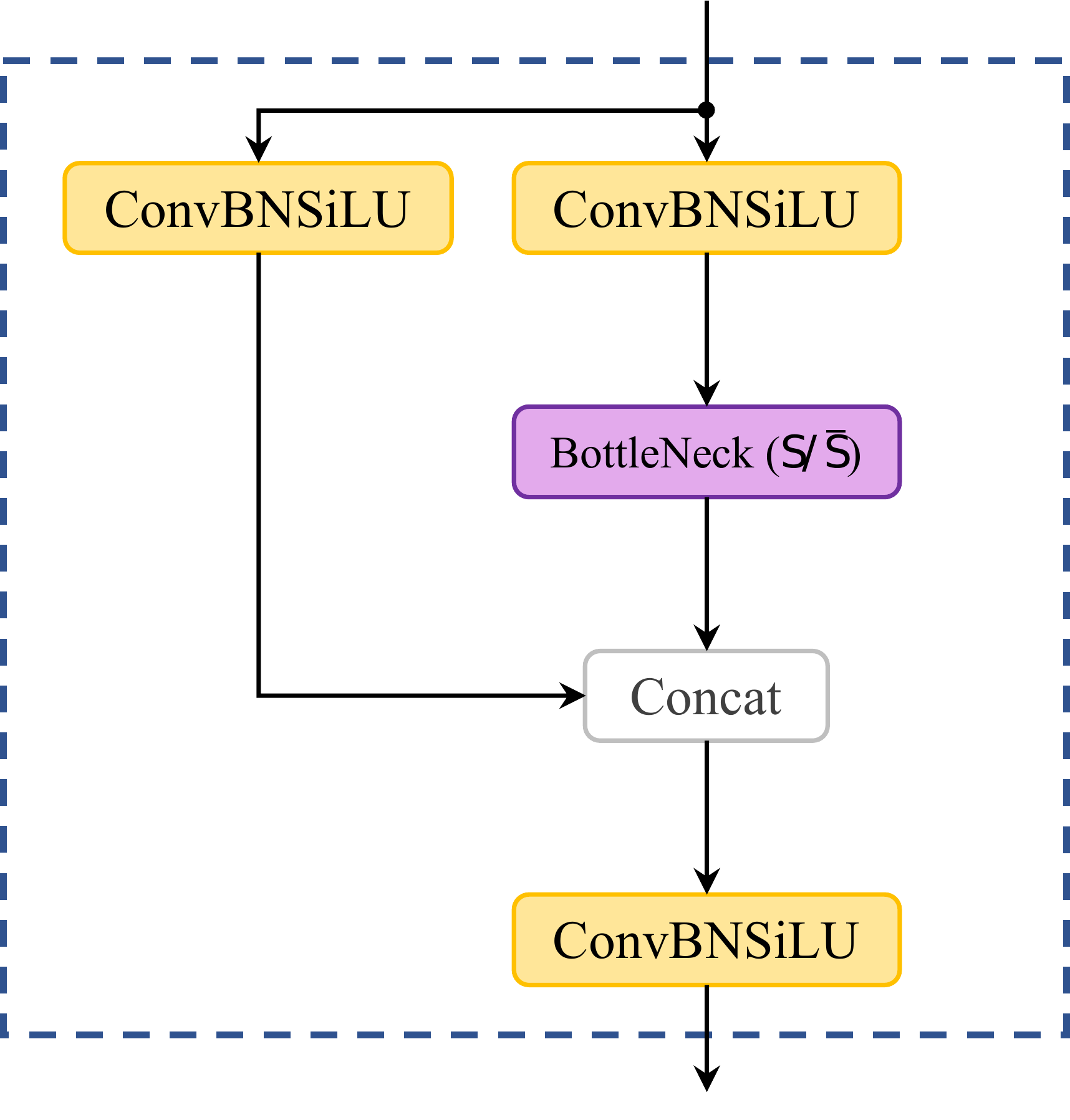}
        \caption{C3 structure}
        \label{fig:C3}
    \end{subfigure}
    \hfill
    \begin{subfigure}[b]{0.32\textwidth}
        \includegraphics[width=\linewidth]{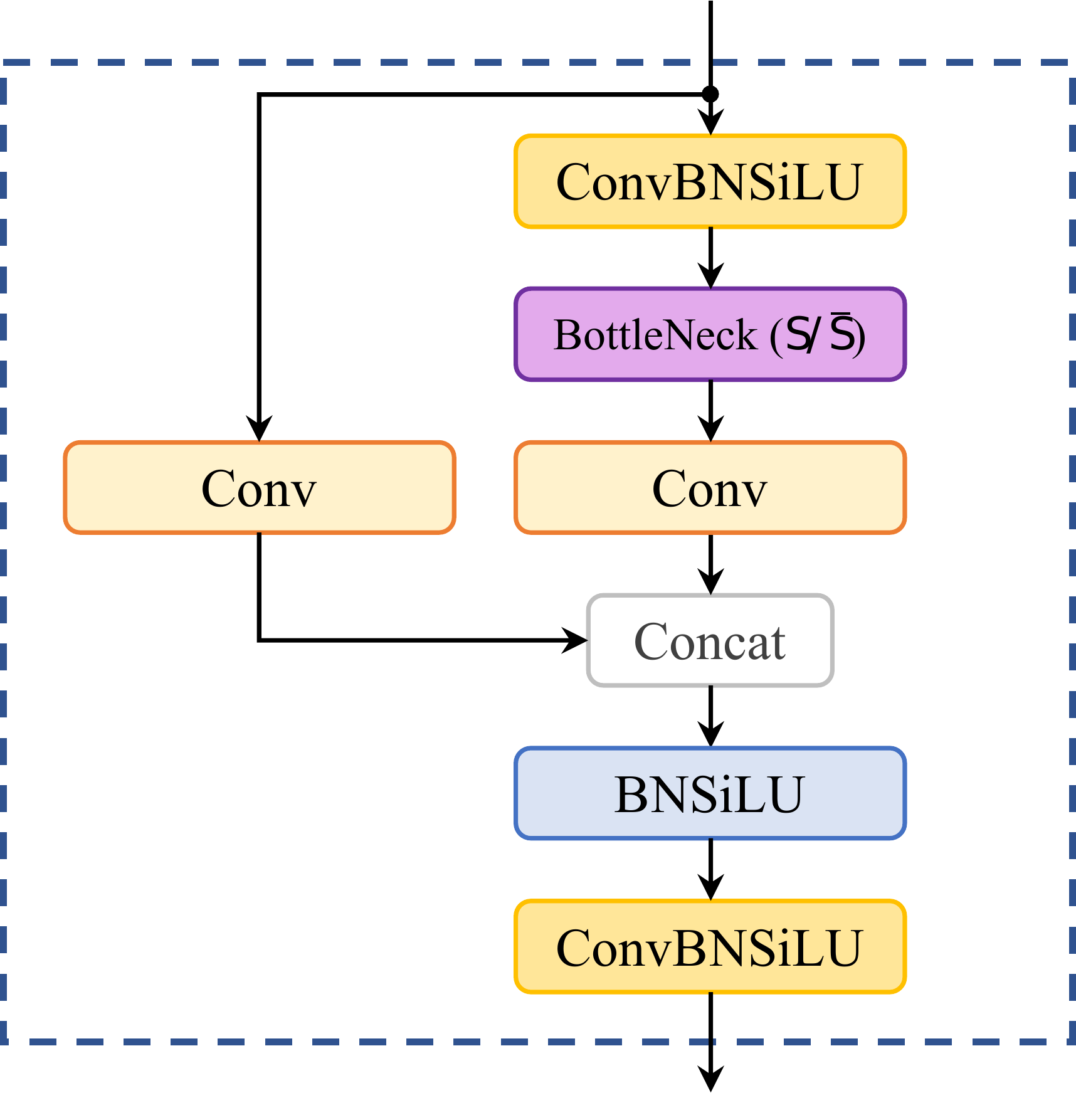}
        \caption{BottleNeckCSP structure}
        \label{fig:BottleNeckCSP}
    \end{subfigure}
    \hfill
    \begin{subfigure}[b]{0.3\textwidth}
        \includegraphics[width=\linewidth]{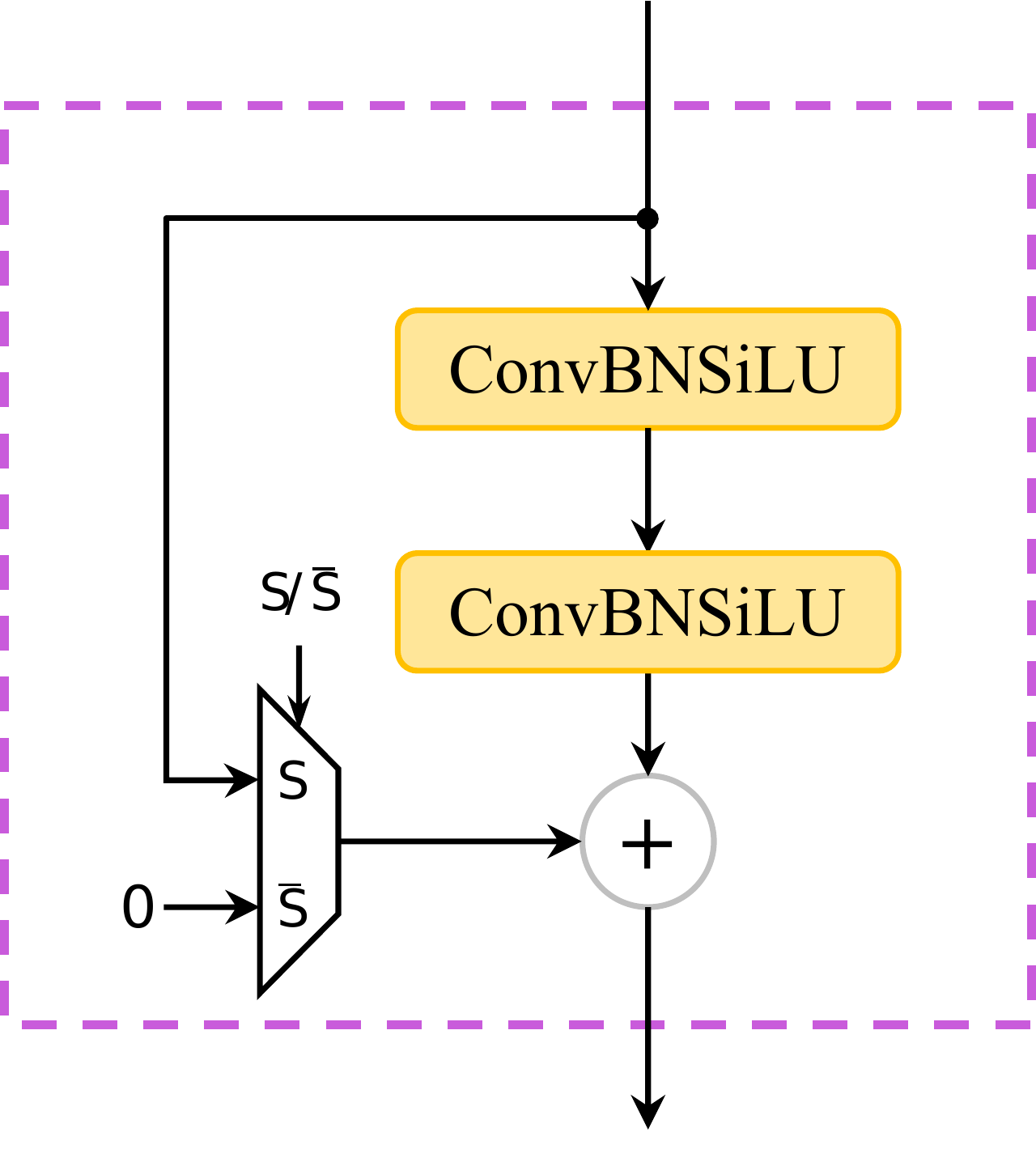}
        \caption{BottleNeck structure}
        \label{fig:BottleNeck}
    \end{subfigure}
    \caption{\textbf{Structure of C3 and BottleNeckCSP modules}. Using the CSP strategy enables the C3 module to strengthen information flow with residual and dense blocks while addressing redundant gradients. BottleNeck block, which is utilized in C3 and BottleNeckCSP and marked as purple, can have two configurations; $S/\bar{S}$. $S$ denotes the active shortcut connection, while $\bar{S}$ characterizes a simple BottleNeck without any skip connection. C3 blocks of the Backbone use BottleNecks with shortcut connection, whereas those of the Neck does not.}
    \label{fig:C3_strc}
\end{figure*}

In 2020, Ultralytics introduced YOLOv5 in five different sizes ranging from nano to extra large \cite{Ultralytics2020YOLOv5}. YOLO underwent major modifications ranging from new backbone architecture to automated hyper-parameter optimization.
In the Backbone, YOLOv5 utilizes a new CSPDarknet53 structure \cite{wang2020cspnet} which is constructed based on Darknet53 with added Cross Stage Partial (CSP) strategy. The design of YOLOv5's Neck takes advantage of CSP-PAN \cite{liu2018path} and a faster variation of SPP block (SPPF). The output is generated through the Head that uses the YOLOv3's Head structure. 
The structure of YOLOv5l is illustrated in Figure \ref{fig:YOLOv5l_whole} in which CSPDarknet53 contains C3 blocks which are the CSP-fused modules.
CSP strategy partitions the feature map of the base layer into two parts and then merges them through a cross-stage hierarchy. Therefore, the C3 module can effectively handle redundant gradients while improving the efficiency of information transfer within residual and dense blocks. C3 is a simplified version of  BottleNeckCSP and is currently used in the latest variant of YOLOv5. For comparison, the design of C3 and BottleNeckCSP blocks are depicted in Figure \ref{fig:C3_strc}.
Overall, these modifications have enabled YOLOv5 to achieve state-of-the-art performance on several benchmarks for object detection, including the COCO dataset. Also, different model size provides the user with the opportunity to choose based on their need.
In 2022, YOLOv6 was released by Meituan, featuring enhancements led by Bi-directional Concatenation (BiC) module, an anchor-aided training (AAT) strategy, and a new backbone and neck design \cite{li2022yolov6}. After a very short period, introduced by the original authors, YOLOv7 \cite{wang2022yolov7} was a breakthrough. Wang et al. proposed a Bag-of-Freebies, a compound model scaling method, and an extended ELAN architecture to expand, shuffle, and merge cardinality. The Bag-of-Freebies refers to a planned re-parameterized convolution inspired by ResConv\cite{ding2021repConv}, an extra auxiliary head in the middle layers of the network for Deep Supervision \cite{lee2015deepsupervision}, and a soft label assigner guiding both the auxiliary head and the lead head by the lead head prediction. Lastly, Ultralytics presented YOLOv8 in 2023 with several alternations in the backbone, neck, and head \cite{YOLOv8}; A C2f module is used instead of C3; a decoupled head provides the output; and the model directly predicts the center of the object instead of the anchor box.
While YOLOv6/7/8 are more featured models, our work focuses on YOLOv5 as more well-established studies have been done on it. However, this investigation can be expanded to newer versions of YOLO, especially YOLOv8.

The current trend in using and expanding over-parameterized models leads to higher accuracy; however, the required Floating-Point Operations (FLOPs) and parameters are dramatically increasing \cite{Mi2022effCNN_survey}. This issue impedes the deployment of complex models on edge devices owing to memory, power, and computational power limitations.
To address this matter, one can employ Cloud Computing (CC). However, running complex models on cloud services may not be a feasible option due to 1) the cost of the network: transferring image data to the cloud consumes relatively large network bandwidth; 2) Latency in time-critical tasks: access delay to cloud services is not guaranteed; 3) Accessibility: cloud services rely on the access of devices to wireless communications which can be disrupted in many environmental circumstances \cite{chen2019DLedge_review}.
Hence, in many cases, edge computing emerges as a more fruitful resolution. Accordingly, various methods have been introduced to compress neural networks with the purpose of making large models deployable on edge devices. Model compression methods can fall under three categories; pruning, quantization, and knowledge distillation. In pruning, redundant parameters of a model with less importance are removed to obtain a sparse/compact model structure. Quantization involves representing models' activations and weights using lower-precision data types. Finally, knowledge distillation refers to employing a large and accurate model as a teacher to train a small model using soft labels supplied by the teacher model(s) \cite{hinton2015distilling, gholami2021sQuanturvey}.

In this review paper, our focus is on pruning and quantization methods since they are widely used as modular compression techniques while utilizing knowledge distillation requires having two models or modifying the structure of the target network. We review the pruning and quantization methods that have been utilized to compress YOLOv5 in recent years and compare the results in terms of compression terminology.  We choose to focus on YOLOv5 as it is the most recent YOLOv5 with enough research on it associated with pruning and quantization. 
Although the newer versions of YOLO have recently outperformed YOLOv5 in many areas, their applied compression methods are still insufficient to review.
Many reviews have been done on neural network compression methods; however, here, we review the actual implementations of such methods on YOLOv5 in real-world situations. We present all the work related to pruning and quantizing YOLOv5, along with their result from different aspects.
Generally, the compression results can be expressed regarding changes in memory footprint, power usage, FLOPs, inference time, FPS, accuracy, and training time.

The paper is organized as follows:
In section \ref{Pruning}, various types of pruning techniques are reviewed, and the recent practical research on applying pruning to YOLOv5 is analyzed. Similarly, in section \ref{Quantization} quantization techniques, along with their empirical implementations on YOLOv5, are reviewed. Finally, in section \ref{FutDirection}, the challenges of compressing YOLOv5 through pruning and quantization are discussed, and possible future directions are provided based on the current gap in this area.

\section{Pruning} \label{Pruning}
Neural network pruning was initially proposed in Optimal Brain Damage \cite{lecun1989optimalDamage} and Optimal Brain Surgeon \cite{hassibi1993optimalSurgeon}. They both rely on a second-order Taylor expansion to estimate parameter importance for pruning. That is, the Hessian matrix should be partially or completely computed in the mentioned methods. However, other criteria can be employed to identify the importance, also called \textit{saliency}, of parameters.
Hypothetically, while the best criterion would be an exact evaluation of each parameter's effect on the network, such evaluation is excessively costly to compute. Therefore, other evaluations, including $\ell_{n}$-norm, the mean or standard deviation of feature map activation, batch normalization scaling factor, first-order derivative, and mutual information, can be utilized for saliency analysis. In the following section, we will discuss such saliency evaluation methods. We will not quantify the effectiveness of each scheme here because different works can hardly be compared, and various factors ranging from hyperparameters to learning rate schedules to architecture of implementation impact the results. Instead, we will present the idea behind each criterion and express the results of applying them to compress YOLOv5.
\subsection{Saliency Criteria for Pruning}\label{section:Saliency}
Saliency criteria refer to measures or metrics used to determine the importance or relevance of individual weights, neurons, filters, or a group of weights in a neural network based on certain characteristics or properties of the network. We refer the reader to \cite{Molchanov2016saliency} for a detailed review of saliency criteria.
\subsubsection{$\ell_{n}$-norm}
Pruning a model based on $\ell_{n}$-norm is the predominantly-used method over the scope of this review paper. Since weight values generally form a normal distribution with zero mean, this is an intuitive approach to select less important individual or structure of weights.
The challenge in using this criterion is to define a threshold with which pruning can be performed. Such a threshold can be set statically for the whole network or for each layer. Also, one can approach it as a dynamic parameter and define schedulers for this threshold. For instance, \cite{softthresh2023unified} proposes a method that treats threshold scheduling as an implicit optimization problem and provides threshold schedulers upon using Iterative Shrinkage-Thresholding Algorithm (ISTA).\\
$\ell_{n}$-norm is usually combined with sparse training of the network to push parameters with the same effect to have similar values \cite{Molchanov2016saliency} (see Section \ref{BNSF}). To do so, $\ell_{1}$ or $\ell_{2}$-regularization is usually added to the cost function, and parameters with a low $\ell_{2}$ norm are pruned after (each step of) training \cite{zhou2016lessmore, memory2014bounded}.

\subsubsection{Feature map activation}
When an activation function is used at the end of a layer, its output can be interpreted as the importance of parameters on the prediction. For instance, in case of ReLU function, outputs closer to zero can be considered less salient and be chosen as candidates for pruning. Also, in a broader outlook, the mean or the standard deviation of a tensor of activation can indicate saliency \cite{Molchanov2016saliency, General2020saliency}.

\subsubsection{Batch normalization scaling factor (BNSF)} \label{BNSF}
Although it can be categorized as a fusion between $\ell_{1}$-norm and feature map activation criteria, the BN scaling factor is used predominantly for pruning YOLOv5 and, more generally, for CNNs. Presented by \cite{liu2017BNFactor}, this method introduces a scaling factor $\gamma$ for each channel and penalizes it during training to obtain a sparse network that can be pruned. The authors proposed the BN scaling factor as the $\gamma$ needed for network compression. In their method, they penalize $\gamma$ of the channels using $\ell_{1}$-norm and then prune channels with near-zero scaling factors.

\subsubsection{First-order derivative}
Unlike the previous criterion, the first-order derivative metrics use the information provided during backpropagation via gradients. This class of criteria may combine information from activations to gradients as expressed in \cite{Molchanov2016saliency, consice2020saliency}.

\subsubsection{Mutual information}
The Mutual Information (MI) between the parameters of a layer and the prediction or that of another layer can represent saliency. In \cite{saliency2018mutual}, authors tried to minimize the MI between two hidden layers while maximizing it between the last hidden layer and the prediction.

\subsection{Granularity of Pruning}
The granularity of pruning defines what kind of parameters of the model are to be pruned. Broadly, pruning can be done either in a structured or unstructured way.

\subsubsection{Unstructured pruning}
Unstructured or fine-grained pruning refers to when goal parameters for pruning are weights of the model without considering their location in the associated tensor or layer. In weight pruning, unnecessary weights are identified through saliency evaluation and masked out or removed afterward. Since eliminating the weights may impair the model architecture, they are mostly masked out during this process \cite{General2020saliency}. While masking out the weights instead of removing them increases the memory usage during training, the information of the masked weights can be used at each step to compare the pruned model with the original one.
fine-grained pruning is not always beneficial as it requires special hardware to take advantage of such irregular sparse operations \cite{nonsprun2020autocompress}.
While higher compression ratios could be achieved using unstructured pruning, storing the index of pruned weights may result in higher storage usage \cite{hdwEffect2021prun, great2021sparsity}.

\subsubsection{Structured pruning}
\label{pruning:structured}
\begin{figure*}[htbp]
  \centering
  \begin{subfigure}[b]{0.45\linewidth}
    \centering
    \includegraphics[width=\linewidth]{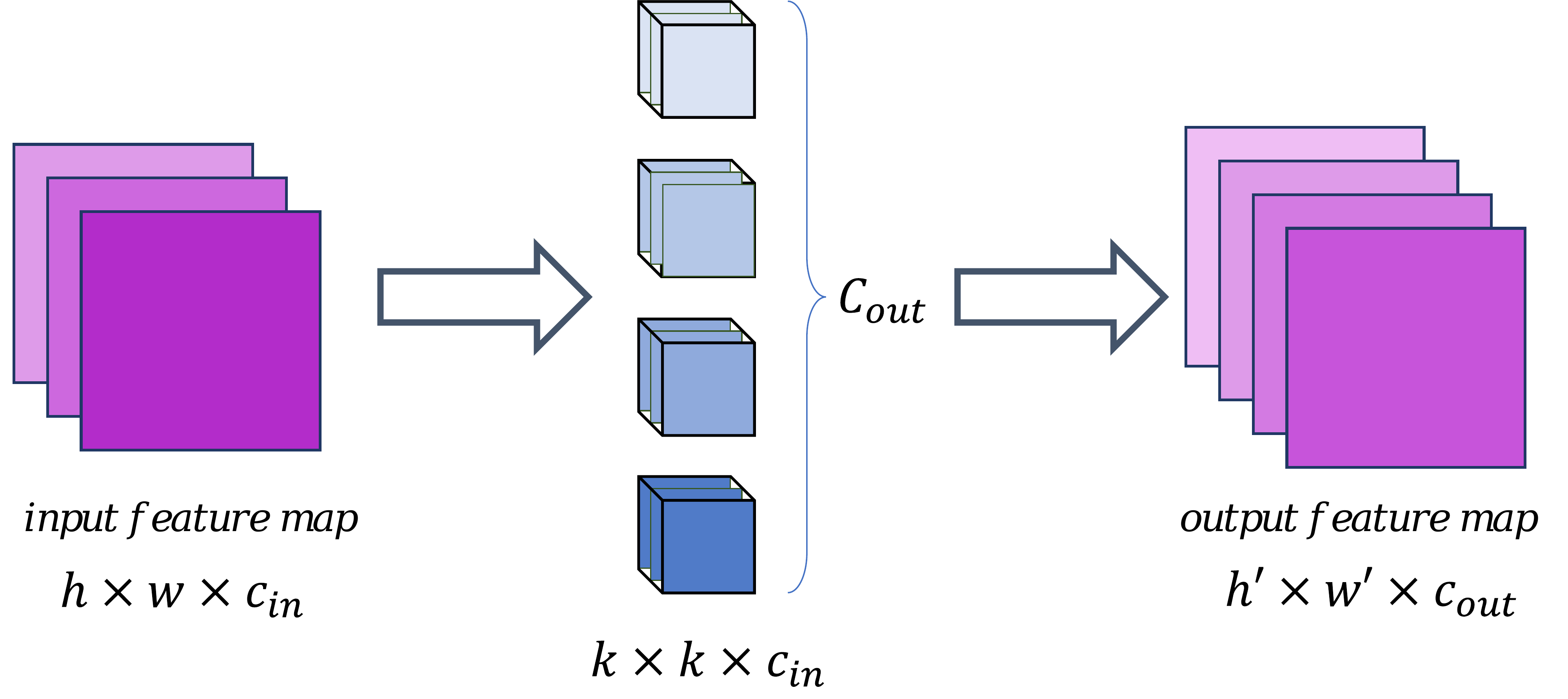}
    \caption{A convolutional layer}
    \label{granul:intro}
  \end{subfigure}
  \hfill
  \begin{subfigure}[b]{0.45\linewidth}
    \centering
    \includegraphics[width=\linewidth]{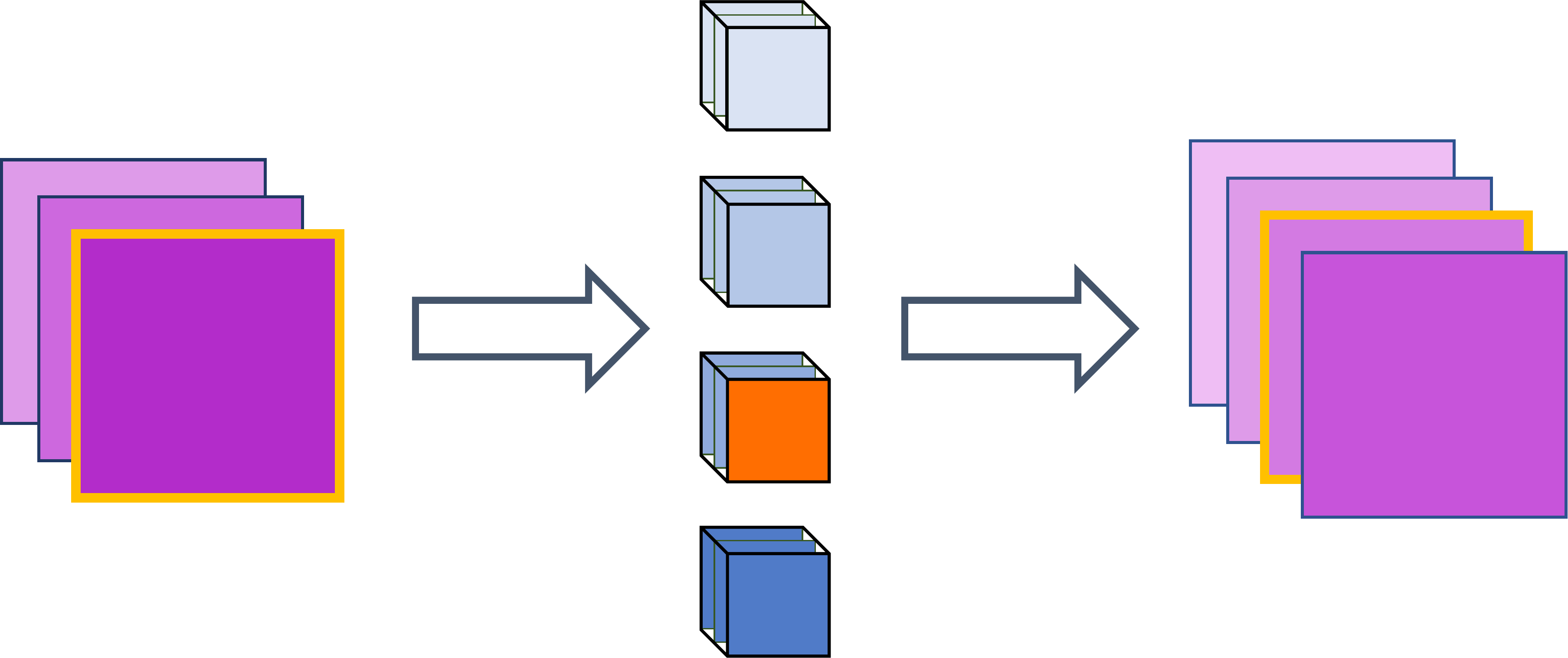}
    \caption{Kernel pruning}
    \label{granul:kernel}
  \end{subfigure}
  
  \vspace{0.5cm}
  
  \begin{subfigure}[b]{0.45\linewidth}
    \centering
    \includegraphics[width=\linewidth]{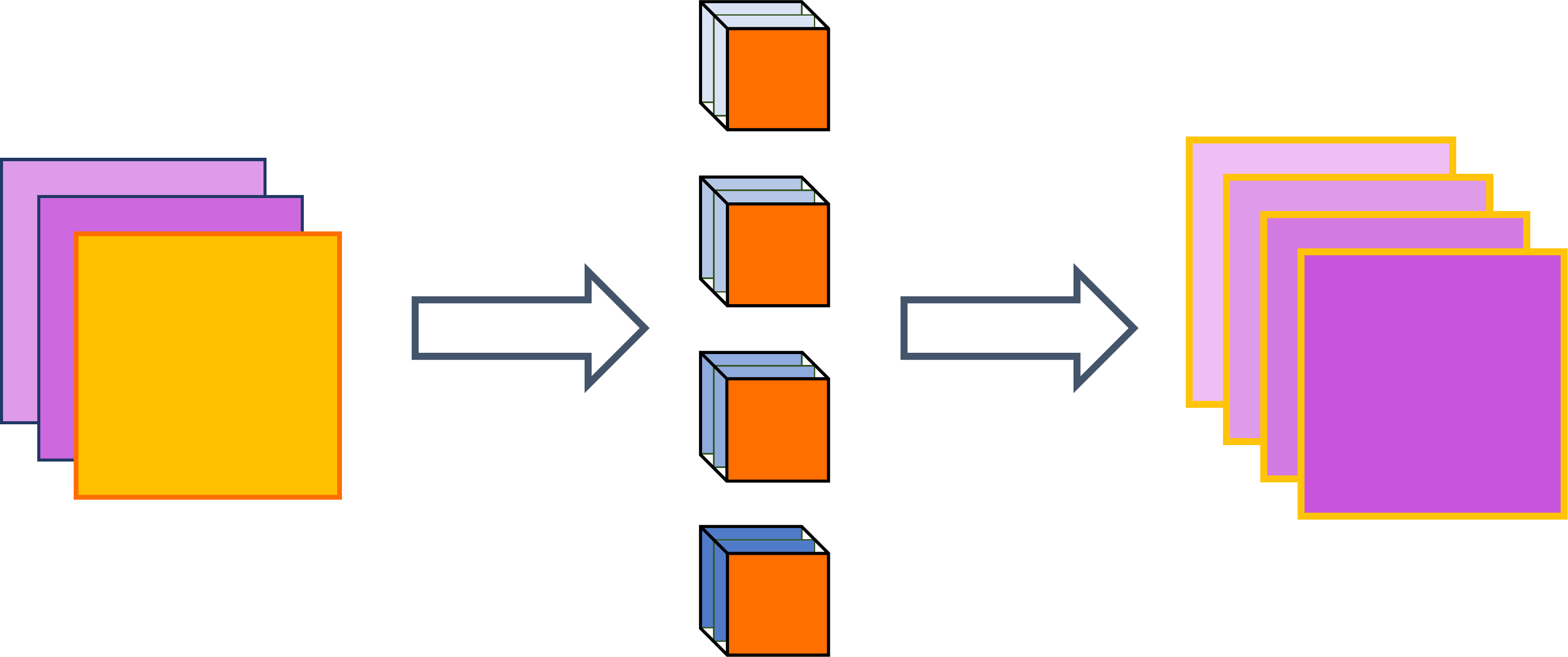}
    \caption{Filter-based pruning}
    \label{granul:filter}
  \end{subfigure}
  \hfill
  \begin{subfigure}[b]{0.45\linewidth}
    \centering
    \includegraphics[width=\linewidth]{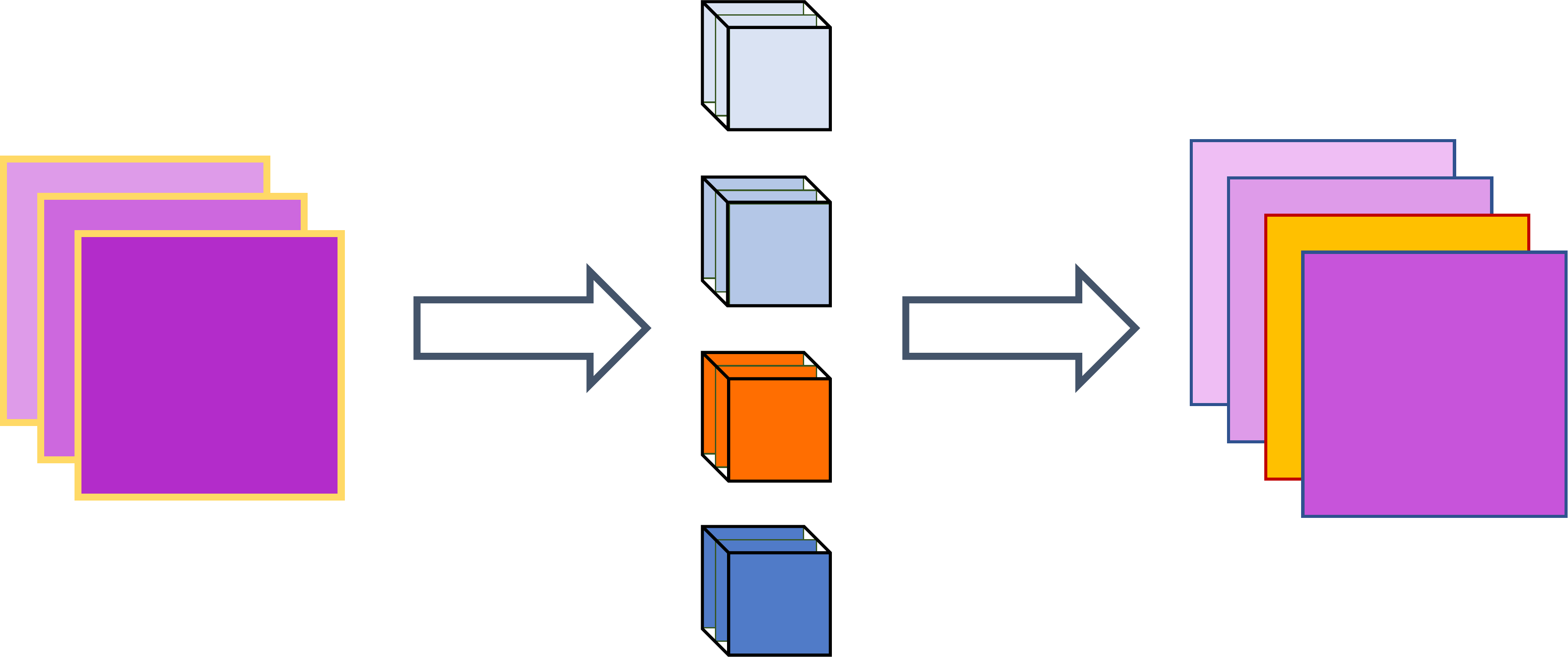}
    \caption{Channel-based pruning}
    \label{granul:channel}
  \end{subfigure}
  \caption{\textbf{Pruning granularity.} The dimension of a convolutional layer is shown in (\subref{granul:intro}). Kernel, Filter, and Channel-based pruning are illustrated in (\subref{granul:kernel}), (\subref{granul:filter}), and (\subref{granul:channel}), respectively. The orange plane characterizes the pruned parameters, and the yellow bounding indicates the affected features.}
  \label{fig:granul:entire}
\end{figure*}

Unlike the previous category, model weights can be pruned based on their structure.
Structured pruning observes patterns in the weights tensors when evaluating their importance so that they can be described with low indexing overhead, such as strides or blocks.
In a convolutional layer, \textit{j}-th channel is obtained through convolving the \textit{j}-th filter with the input feature map. So groups of parameters such as filters, channels, or kernels can be selected for structured pruning. Figure \ref{fig:granul:entire} depicts the difference between these structural pruning paradigms.

\textit{Channel-based pruning.}
It aims at removing the weight filters leading to the \textit{j}-th channel(s) of the output feature map in each layer. Many existing channel pruning techniques utilize the $\ell_2$-norm as the criterion for identifying the least important tensor of weights. However, there is a dispute regarding the impact of this process on the overall model structure. In the \cite{iccv2017CHpruning}, the authors stated that the process of channel pruning resulted in little damage to the model structure. Conversely, in the \cite{channel2020pruning}, it was observed that channel pruning led to a drastic change in the network structure.
Nonetheless, it is possible to mitigate structural damage by masking the parameters rather than completely eliminating them. However, this approach may not result in any savings during training since the entire model needs to be stored in memory.

\textit{Filter-based pruning.} Filter-based pruning eliminates the weights corresponding to the \textit{i}-th channel of the input feature map. That is, pruning a specific kernel, \textit{i}-th, of all filters in a convolutional layer. The structural damage is minimal in this pruning method, and the model can be treated similarly to the original one since the number of output channels remains intact \cite{Mi2022effCNN_survey}.  It is worth mentioning that: 1) channel-based pruning at \textit{l}-th layer is equivalent to filter-based pruning at \textit{(l+1)}th layer, and 2) Filter pruning is not equivalent to filter-based pruning. In filter pruning, one or some of the filters of a layer are pruned which can be categorized as channel-based pruning from the granularity perspective.

\textit{Kernel-based pruning.} In this category, all parameters over a kernel of a filter in \textit{l}-th layer, which connects the \textit{i}-channel of the input feature map and \textit{j}-th channel of the output feature map, are pruned \cite{forKernel2017}. This pruning granularity would not impair the model structure. We refer the interested reader to \cite{Mi2022effCNN_survey} for a thorough survey of pruning granularity.\\
Regardless of the pruning granularity and the saliency criteria, the pruning process can be performed either in a one-shot manner or iteratively. In one-shot pruning \cite{oneshot2021alg, oneshot2022exm, oneshotCH2019alg}, less salient parameters are removed/masked prior to or after training. In post-training pruning, network performance may drop permanently, while iterative pruning, takes the performance drop into consideration and re-trains the network. Compared to one-shot pruning, although iterative pruning is more computation- and time-consuming, it can prevent the accuracy drop or even enhance it in some cases. Moreover, some methods modify the network cost function, such as adding regularization \cite{liu2017BNFactor, shao2022structured}, to make the model more suitable for pruning. Consequently, they cannot be used as post-training pruning.\\ 

\begin{landscape}
\begin{table*}[tb]
\scriptsize
    \caption{Experimental results of pruning applied on YOLOv5 categorized by granularity. Each granularity method is specified by a color. $\Delta$Acc represents the absolute difference in $mAP^{0.5}$ before and after pruning. The rest of the quantitative results represent relative changes. Also, \rightturn~denotes the iterative pruning, \astrosun~denotes the one-shot pruning, and $\Delta$\clock~refers to the changes in the inference time.}
    \resizebox{\linewidth}{!}{
    \begin{tabular}{C{1.1cm}||C{0.5cm}C{4cm}C{2cm}|C{1.6cm}C{1.5cm}C{0.5cm}|C{0.8cm}C{0.8cm}C{0.8cm}C{0.8cm}C{1.6cm}}
    \hline
        \textbf{Paper} & \textbf{V} & \textbf{Task} & \textbf{Dataset} & \textbf{Saliency} & \textbf{Granul} & \textbf{\rightturn/\astrosun} & \textbf{$\Delta$Acc} & \textbf{$\Delta$Param} & \textbf{$\Delta$Size} & \textbf{$\Delta$FLOPs} & \textbf{$\Delta$FPS}/\textbf{$\Delta$\clock} \\ \hline
        
        \multicolumn{10}{l}{\textbf{Unstructured Pruning}} \vspace{.1cm} \\
        \rowcolor{ORANGE} \cite{frantar2022spdyUnstructured} & 5s & Object Detection & --- & SPDY & unstructured & \rightturn & -0.5 & --- & --- & --- & 50/--- \\ 
        \rowcolor{ORANGE} \cite{frantar2022spdyUnstructured} & 5m & Object Detection & --- & SPDY & unstructured & \rightturn & -1.7 & --- & --- & --- & 75/---~ \\ \hline
        \multicolumn{10}{l}{\textbf{Channel-Based Pruning}} \vspace{.1cm} \\
        \rowcolor{CHANNEL} \cite{pan2023lightweight} & 5s & Position Detection & Manual & BNSF & channel & \rightturn & -2.3 & -45 & -48 & -55 & ~~---/-40 \\ 
        \rowcolor{CHANNEL} \cite{yang2022new} & 5n & Fault Detection & Manual & BNSF & channel & \rightturn & -4.8 & -72.2 & --- & --- & ~~~~---/-29.4 \\ 
        \rowcolor{CHANNEL} \cite{chen2022improved} & 5 & Target Detection & Military Aircraft Detection & BNSF & channel & \rightturn & -4.5 & --- & --- & --- & 560/---~~~ \\
        \rowcolor{CHANNEL} \cite{wang2021channel} & 5s & Fruitlet Detection & Manual & BNSF & channel & \rightturn & 0(F1) & -92 & -90.5 & --- & ~~---/-13 \\ 
        \rowcolor{CHANNEL} \cite{li2022channel} & 5 & Outdoor Obstacles Detection & OBSTACLE & BNSF & channel & \rightturn & -0.4 & --- & -59.1 & --- & ~~~~---/-43.6 \\ 
        \rowcolor{CHANNEL} \cite{ding2022design} & 5s & Garlic Detection & Manual & BNSF & channel & \rightturn & -0.1 & --- & -17 & --- & ---/-1 \\ 
        \rowcolor{CHANNEL} \cite{zhao2022fast} & 5s & Wheat Grain Quality Detection & Manual & --- & channel & \rightturn & 1 & --- & --- & -86.7 & ---/--- \\ 
        \rowcolor{CHANNEL} \cite{zheng2022fast} & 5x & Ship Detection & Manual & BNSF & channel & \rightturn & 2.3 & --- & -48.9 & --- & 61.4/---~~~~ \\ 
        \rowcolor{CHANNEL} \cite{zhou2021flame} & 5s & Flame Detection & Manual & BNSF & channel & \rightturn & 0 & -37.8 & -37.5 & -54.5 & 10/---~~ \\ 
        \rowcolor{CHANNEL} \cite{li2022improved} & 5s & Satellite Components Recognition & Manual & BNSF & channel & \rightturn & -1.2 & --- & -66.2 & --- & ---/--- \\ 
        \rowcolor{CHANNEL} \cite{sun2022mca} & 5s & Helmet-Wearing Detection & --- & BNSF & channel & \rightturn & -0.9 & -87.3 & -86.2 & -87.4 & 53.5/---~~~~ \\ 
        \rowcolor{CHANNEL} \cite{zhang2021pruned} & 5l & Object Detection & COCO & CDSC-BNSF & channel & \rightturn & -0.9 & --- & -63.8 & -37.4 & ---/--- \\ 
        \rowcolor{CHANNEL} \cite{situ2023real} & 5s & Sewer Defect Detection & Manual & BNSF & channel & \rightturn & -0.5 & -81 & -79.3 & -48.8 & ~~~~---/-34.2 \\ 
        \rowcolor{CHANNEL} \cite{shen2023real} & 5s & Tracking and Counting Grape Clusters & Manual & BNSF & channel & \rightturn & -0.2 & -73.3 & -76.4 & -57.6 & ~~~~---/-10.3\\ 
        \rowcolor{CHANNEL} \cite{zhang2022research} & 5 & Fiber Defect Detection & PASCAL VOC & BNSF & channel & \rightturn & -0.3 & --- & -29.5 & --- & ~~~~---/-18.7 \\ 
        \rowcolor{CHANNEL} \cite{liu2022research} & 5s & Pedestrian Detection & VisDrone & BNSF & channel & \rightturn & -0.4 & --- & -26.3 & -11.9 & ~~~---/-9.3 \\ 
        \rowcolor{CHANNEL} \cite{zhang2022sod} & 5 & Blade Defect Detection & Manual & BNSF & channel & \rightturn & 7.8 & --- & -19.6 & --- & 28.3/---~~~~ \\ 
        \rowcolor{CHANNEL} \cite{jeon2022target} & 5l & Object Detection & PASCAL VOC & CRLST & channel & \rightturn & -2.9 & -49.2 & --- & -46.2 & ~~~~---/-35.5 \\ 
        \rowcolor{CHANNEL} \cite{jeon2022target} & 5l & Object Detection & COCO & CRLST & channel & \rightturn & -3.8 & -48.8 & --- & -46.5 & ~~~~---/-29.4 \\ 
        \rowcolor{CHANNEL} \cite{jeon2022target} & 5l & Object Detection & VisDrone & CRLST & channel & \rightturn & -1.1 & -51.2 & --- & -46 & ~~~~---/-35.4 \\ 
        \rowcolor{CHANNEL} \cite{lv2022yolov5} & 5s & Pedestrian Detection & Manual & BNSF & channel & \rightturn & 3.78 & -52.3 & -51.8 & -40.8 & 57.8/---~~~~\\
        \rowcolor{CHANNEL} \cite{xu2022compressed} & 5s & Aerial Image Object Detection & DOTA & BNSF & channel & \astrosun & -6.46 & -58.8 & -57.6 & -37 & ~~~~---/-17.3 \\
        \rowcolor{CHANNEL} \cite{nan2023faster} & 5l & Object Detection & Manual & NSGA-II & channel & \rightturn & -0.9 & -73.9 & -73.5 & -62.7 & 59/---~~ \\
        \rowcolor{CHANNEL} \cite{shao2022structured} & 5m & Object Detection & VisDrone & BNSF \& ASR & channel & \rightturn & 0.4 & -91.2 & --- & -65.8 & ~~~~---/-65.7 \\
        \rowcolor{CHANNEL} \cite{shao2022structured} & 5s & Object Detection & VisDrone & BNSF \& ASR & channel & \rightturn & -0.5 & -93.4 & --- & -74.3 & ~~~~---/-53.2 \\ 
        \rowcolor{CHANNEL} \cite{zeng2023lightweight} & 5 & Tomato Maturity Detection & Manual & --- & channel & --- & -1.5 & -78 & -76.4 & -84.1 & 183/-64.9 \\
        \rowcolor{CHANNEL} \cite{wang2022defect} & 5s & Railway Defect Detection & Manual & FPGM & channel & \rightturn & 2.19 & --- & --- & --- & 34.7/---~~~~ \\ \hline
        
        \multicolumn{10}{l}{\textbf{Hybrid Pruning}} \vspace{.1cm} \\
        \rowcolor{HYBRID} \cite{song2021object} & 5l & Object Detection & Manual & --- & kernel/layer & --- & 1.27 & --- & -86.8 & --- & ~~~~---/-58.4 \\ 
        \rowcolor{HYBRID} \cite{wang2022apple} & 5s & Fruit Detection & Manual & BNSF & channel/layer & \rightturn & -1.57 & --- & -72.5 & --- & 32.5/---~~~~ \\ \hline
    \end{tabular}}
    \label{table:pruningGranul}
\end{table*}
\end{landscape}

\subsection{Recent Applied Studies on Pruned YOlOv5}

Table \ref{table:pruningGranul} represents the recent experimental pruning results on YOLOv5 categorized by the pruning granularity. \cite{frantar2022spdyUnstructured} concentrates on achieving a desired inference time instead of a specific compression ratio. It presents a pruning method, \textit{learned efficient Sparsity Profiles via DYnamic programming search} (SPDY), that can be utilized in both one-shot and iterative schemes. In \cite{nan2023faster}, researchers implemented an algorithm based on Non-Dominated Sorting Genetic Algorithm (NSGA) II that tackles pruning as an optimization problem. That is, how to prune the channels in order to minimize GFLOPs and maximize the $mAP^{0.5}$. \cite{shao2022structured} introduces an 
Adaptive Sparsity Regularization (ASR) that renders sparse constraints based on the filter weights. That is, a penalty is assigned to the filters with weak channel outputs in the regularized loss function, instead of directly regulating the loss with the L1-norm of the batch normalization scaling factor. After training, filters with a scaling factor smaller than a global threshold are pruned for all layers, and find-tuning is performed to retrieve the accuracy drop. The work of \cite{pan2023lightweight} modifies the YOLOv5 structure by utilizing the PReLU activation function and using Ghost Bottleneck instead of BottleNeckCSP. Thereafter, it prunes the channels, except Ghost Bottleneck, with small batch normalization scaling factor based on the BNSF method (see Section \ref{BNSF}). \cite{li2022improved} proposed a feature fusion layer and selective kernel network to improve the channel and kernel attention of the model. It attaches transformer encoder modules to the outputs of the PAN neck to explore the potential of the model's prediction via a self-attention procedure and compresses the model using the BNSF approach before deployment on NVIDIA Jetson Xavier NX. 
Furthermore, \cite{jeon2022target} targets a desired number of parameters and FLOPs and utilizes a computation-regularized loss sparse training (CRLST). After sparse training, it iteratively prunes the channels based on their batch normalization scaling factor. Both compressed YOLOv5 models in \cite{jeon2022target, li2022improved} were deployed on NVIDIA Xavier NX.
\cite{wang2022defect} treats filters as points in space and adopts a Filter Pruning Geometric Median (FPGM) approach to prune filters of a convolutional layer that, unlike $\ell_{n}$-norm criteria, explicitly exploits the mutual relations between filters. It computes the geometric median of a whole layer's weights and prunes the filters which are regarded redundant if their geometric median is close to the geometric median of the layer. \cite{zhang2021pruned} discards the upsample, concatenate, and detect layer and prunes the filters similar to the BNSF method, yet it embeds the cosine decay of sparsity coefficient (CDSC-BNFS) and uses a soft mask strategy. \cite{liu2022research, zhang2022sod} makes the model lighter by BNSF channel-based pruning and adds another upsampling level with the BottleNeckCSP module to the neck network to extract more semantic information from small objects. The latter also adds a Convolution Block Attention Module (CBAM) to the output of each BottleNeckCSP module in the neck network before feeding them to the head. The paper \cite{wang2022apple} considers the mean value of each Conv layer before each shortcut in the BottleNeckCSP and compresses the backbone. It also performs channel-based pruning using the BNSF method. In the following, several recent implementations of pruning on YOLOv5 are listed:

\begin{itemize}
\item \cite{xu2022compressed, yang2022new} prunes the network through the BNSF method but combines the fine-tuning with knowledge distillation to save training time while maintaining accuracy.

\item The Authors in \cite{chen2022improved} replaced the CSPDarknet backbone with MobileNetV3 \cite{howard2019mobilenetv3} and use TensorRT after pruning the filters with the BNSF method. 
 
\item The work of \cite{wang2021channel, li2022channel, ding2022design, zhou2021flame, situ2023real} focuses on pruning the filters using BNSF strategy and fine-tuning.

\item In \cite{zhao2022fast}, the backbone is pruned, and since the desired objects have relatively the same size, the largest feature map of the PAN module is removed. Also, in the neck, a hybrid attention module is proposed to extract the most comprehensive feature from channels.

\item \cite{zheng2022fast} employs the t-Distributed Stochastic Neighbour Embedding algorithm to reduce the dimension of anchor frame prediction and fuses it with weighted clustering to predict frame size to achieve a more accurate prediction target frame. Afterward, it prunes the filters through the BNSF method.

\item \cite{sun2022mca} utilizes a multi-spectral channel attention mechanism in the backbone to generate more informative features and improve the model's accuracy for detecting small objects. Then, using the BNSF process, it prunes the filters of the model.

\item \cite{shen2023real} lightens the model through pruning filters with the BNSF criteria and introduces a soft non-maximum suppression which enables the model to detect overlapping clusters of grapes instead of discarding them.

\item \cite{zhang2022research} combines feature maps from different receptive fields using spatial pyramid dilated convolutions (SPDCs) to integrate defect information at multiple scales. It embeds a channel attention mechanism in the neck to direct more attention to effective feature channels after each concatenation operation. Afterward, it compresses the model via BNSF channel-based pruning with fine-tuning.

\item \cite{lv2022yolov5} makes many modifications to the YOLOv5 structure, such as appending an attention mechanism to the neck and a context extraction model to the backbone. As for pruning, it removes the filters using the BNSF criteria.

\item \cite{song2021object} compress the neck and the backbone of YOLOv5 with layer and kernel pruning.

\item \cite{zeng2023lightweight} replaces the backbone of YOLOv5 with MobileNetV3 and prunes the neck network through channel-based pruning.

\end{itemize}

Almost 85\% of the research on pruning YOLOv5 is done using channel-based pruning, and the rest is associated with other structured and unstructured granularities. The main saliency criterion used for pruning is the BNSF sparse training method which is employed in around 60\% of the surveyed papers in our scope, while the rest employed $\ell_{1}$-norm or $\ell_{2}$-norm or proposed a new saliency criterion. 

\section{Quantization} \label{Quantization}

Neural network quantization aims to represent the weights and activations of a deep neural network with fewer bits than their original precision, which usually is 32-bit single-precision floating-point (FP32). This process is done while the effect on the model's performance/accuracy is kept minimal.
Quantization, by exploiting faster hardware integer instructions, can reduce the size of the model and improve inference time \cite{integerQT2020}. In \cite{gholami2021sQuanturvey}, Gholami et al. surveyed different aspects of neural network quantization, which includes the theoretical details of this topic. Here, we will briefly introduce and discuss the key points.
\\
Without loss of generality, we explain the concepts of quantization on a real finite variable which can represent weights or activations in a neural network.
Assuming $r\in \mathbb{R}$ is a finite variable limited to the range of $\mathbb{S}$, we want to map its values to $q$ with a set of discrete numbers that lie in $\mathbb{D} \subset \mathbb{R}$. Before mapping, we may want to clip the range of input $r$ to a smaller set of $\mathbb{C} \subseteq \mathbb{S}$.

\subsection{Quantization Interval: Uniform and Non-uniform }
\begin{figure*}[tb]
  \centering
  \begin{subfigure}[b]{0.45\textwidth}
    \includegraphics[width=\textwidth]{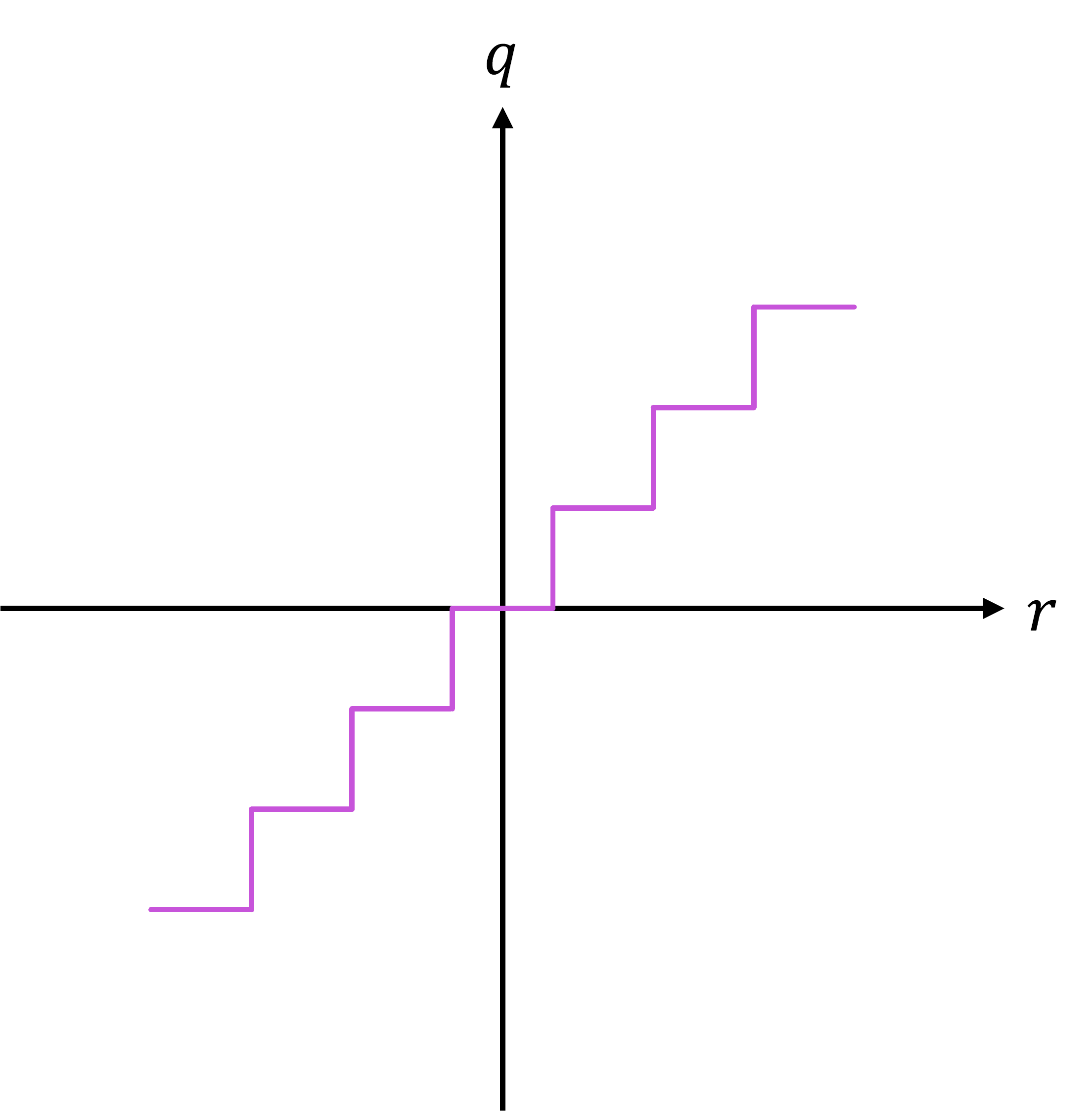}
    \caption{uniform quantization}
    \label{fig:unoform}
  \end{subfigure}
  \hfill
  \begin{subfigure}[b]{0.45\textwidth}
    \includegraphics[width=\textwidth]{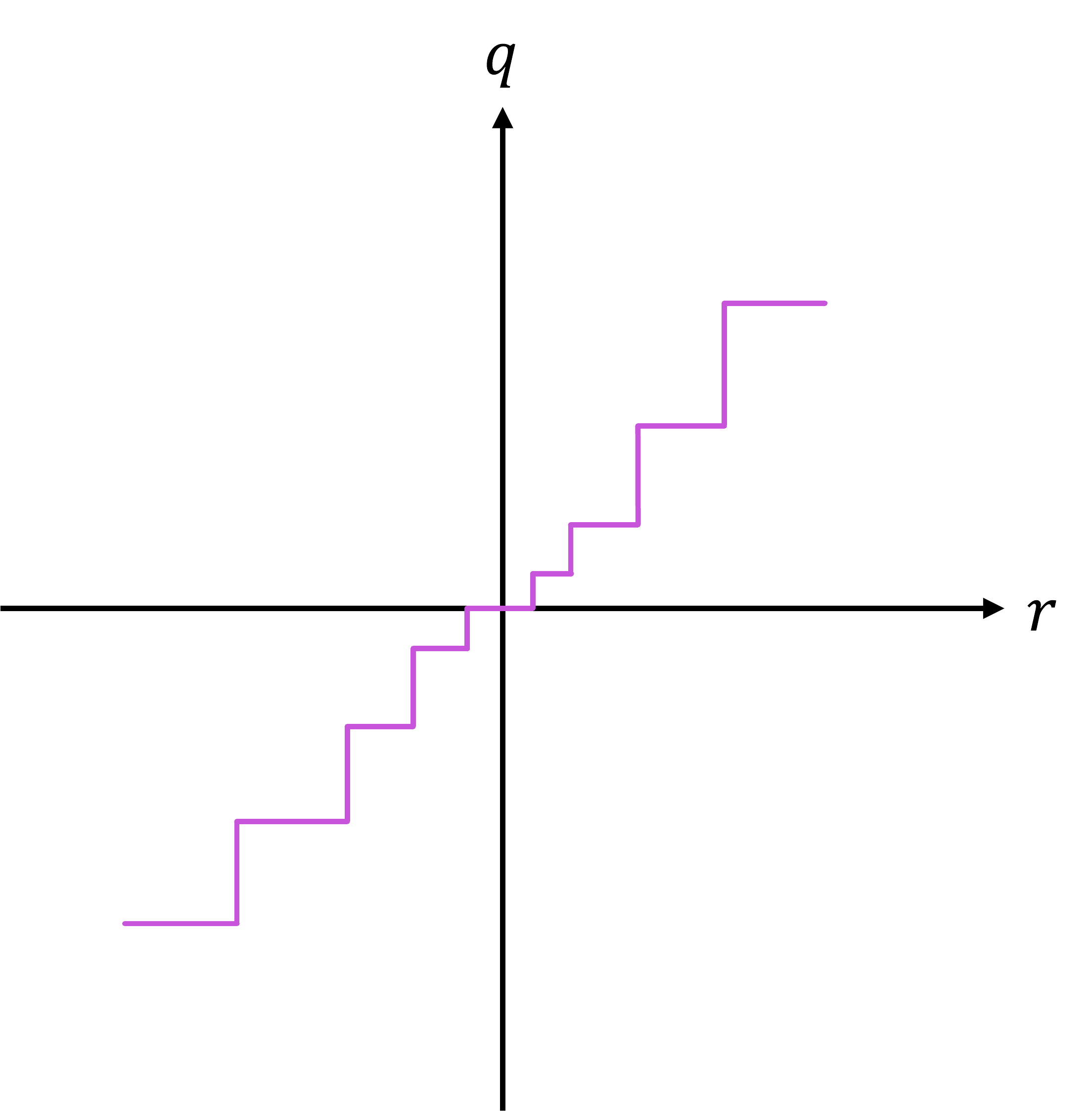}
    \caption{non-uniform quantization}
    \label{fig:nonuniform}
  \end{subfigure}
  \caption{\textbf{Quantization Interval.} (a) indicates the uniform quantization, while (b) depicts the non-uniform quantization in which neither the interval in the real number nor the quantization steps are evenly distributed.}
  \label{fig:entire:QTuniform}
\end{figure*}

Uniform quantization maps $r$ into a set of evenly spaced discrete values, while in non-uniform quantization, the distance between discrete values is not necessarily equal. Through non-uniform quantization, one can better capture the vital information from the weights and activation distributions because, for instance, more closely-spaced steps can be allocated to denser areas of distribution. Consequently, although employing non-uniform quantization requires more design than the uniform approach, it may achieve a lower accuracy drop \cite{ICML2020dtinyscript}. Also, since the distribution for weights and activations generally tends to be bell-shaped with long tails, non-uniform quantization can lead to better results \cite{Li2020, Baskin2021UNIQ, LCQ2021nonuni}. Figure \ref{fig:entire:QTuniform} demonstrates the disparity between the aforementioned quantization schemes.

\subsection{Static and Dynamic Quantization}
For a set of inputs, the clipping range $\mathbb{C}=[a,b]$, where $(a,b)\in \mathbb{R}$, can either be determined dynamically or statically. The former computes the clipping range dynamically for each input, whereas the latter uses a pre-calculated range to clip all the inputs. Dynamic quantization reaches higher accuracy compared to static one, but the computational overhead is significantly high.

\subsection{Quantization Scheme: QAT and PTQ}
Quantizing a trained model can negatively impact the accuracy of the model due to cumulative numerical errors. To recover this drop in performance, network parameters often need adjustment. Therefore, quantization can be performed in two fashions; Quantization Aware Training (QAT), which refers to retraining the network, or Post Training Quantization (PTQ), which does not include re-training.
In QAT, the forward and backward passes of the quantized model are performed in floating points, and network parameters are quantized after each gradient update. 
On the other hand, PTQ performs quantization and parameter adjustment without re-training the network. Compared to QAT, this method usually suffers from the model's accuracy degradation, but its computational overhead is considerably lower. Generally, PTQ uses a small set of calibration data to optimize the quantization parameters and then quantizes the model \cite{PTQ2021Olivia}. As PTQ relies on minimal information, it is often impossible to achieve lower than 4 or 8 bits precision while maintaining accuracy \cite{PTQ2021int4}.\\

\subsection{Quantization deployment scheme}
\begin{figure*}[t]
  \centering
  \begin{subfigure}[b]{0.48\textwidth}
    \centering
    \includegraphics[width=0.7\textwidth]{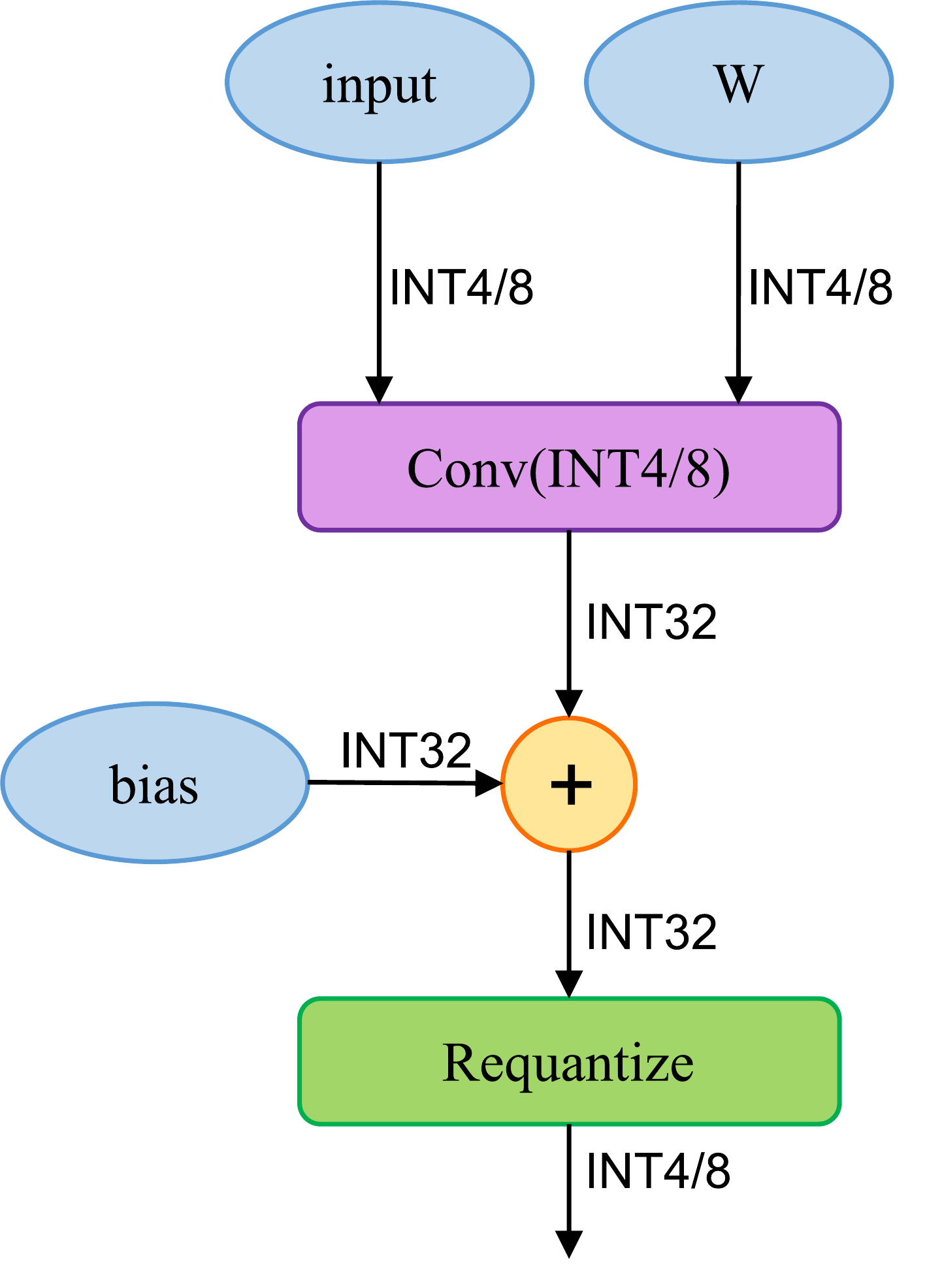}
    \caption{Fixed-precision quantization}
    \label{fig:IntOnly}
  \end{subfigure}
  \hfill
  \begin{subfigure}[b]{0.48\textwidth}
    \centering
    \includegraphics[width=0.7\textwidth]{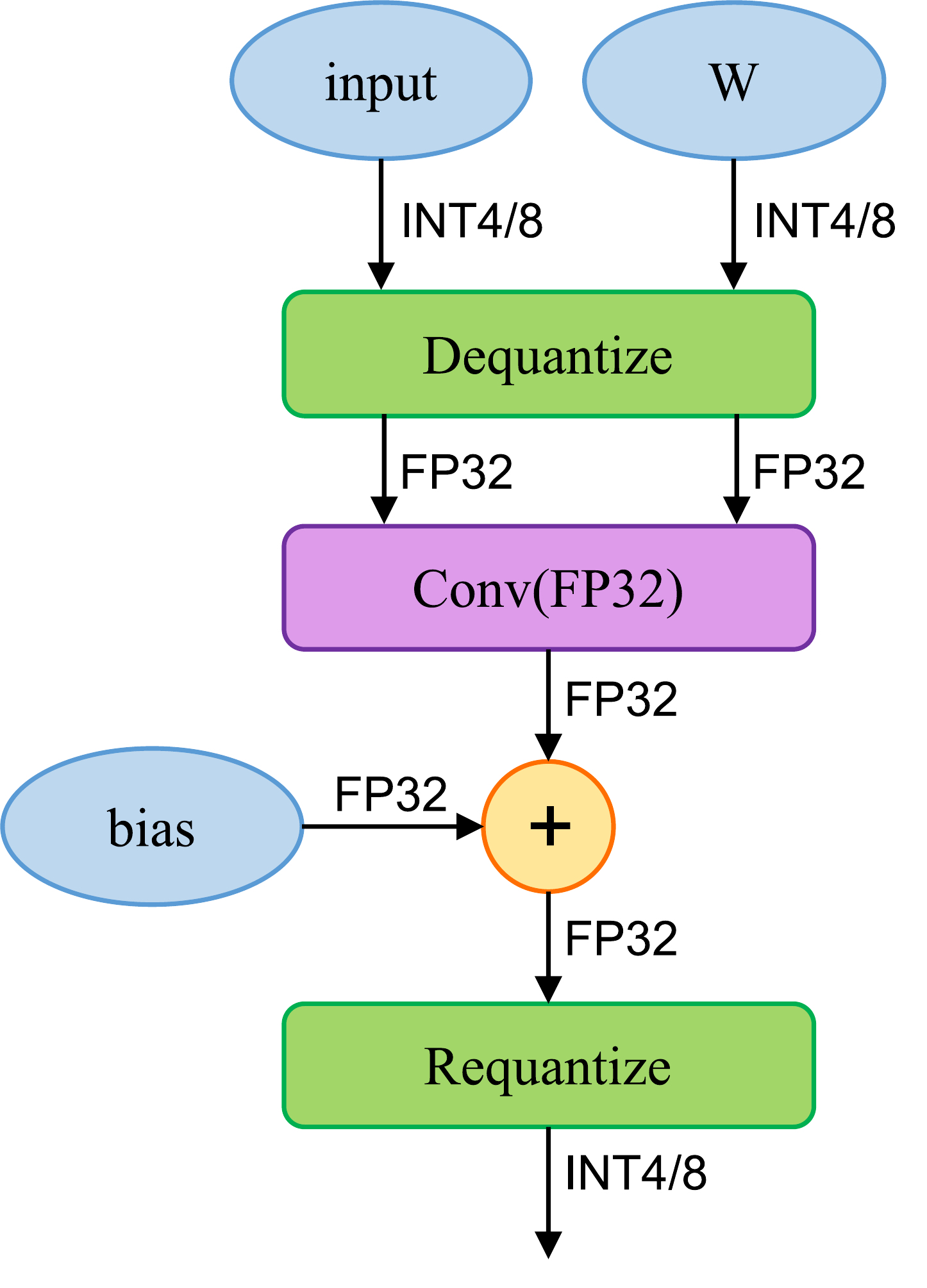}
    \caption{Simulated quantization}
    \label{fig:FakeQuant}
  \end{subfigure}
  \caption{Inference comparison between fixed-precision and simulated quantization in a convolutional layer.} 
  \label{fig:entire:fakeVsFixedQT}
\end{figure*}

Once the model is quantized, it can be deployed using \textit{fake quantization} (also called simulated quantization) or\textit{ integer-only} quantization (also known as fixed-point quantization). In the former, weights and activations are stored in low precision, but all the operations ranging from addition to matrix multiplication are performed in floating point precision. While this approach requires constant dequantizing and quantization before and after floating point operations, it favors the model's accuracy. 
However, in the latter, operations, as well as weights/activation storing, are performed using low-precision integer arithmetic. In such a fashion, the model can take advantage of fast integer arithmetic enabled by most hardware. Figure \ref{fig:entire:fakeVsFixedQT} illustrates the disparity between PTQ and QAT deployment for a single convolutional layer.

\subsection{Recent Applied Studies on Quantized YOLOv5}
Table \ref{table:QuantizationScheme} represents the recent practical quantization results on YOLOv5 categorized by quantization scheme. The Authors in \cite{wu2023drgs}, proposed a QAT method that dynamically selects rounding mode for weights considering the direction of weight updates during training and adjusts the corresponding gradient. They quantized the model layer-wise while adopting symmetric/asymmetric clipping ranges for the weights and activations of the network, respectively.
Noise Injection Pseudo Quantization (NIPQ) \cite{park2022nipq}, as a QAT method, initially pre-trains a network with pseudo quantization noise and then quantizes the model in post-training. This approach automatically adjusts the bit-width and quantization interval while also regularizing the sum of the Hessian trace of a neural network. The authors evaluated their method on YOLOv5 and achieved down to 3-bit precision without major degradation in accuracy. \\
Furthermore, \cite{papaioannou2023ultra} uses ShuffleNetV2 \cite{ma2018shufflenetV2} to modify the backbone and reduce the number of layers in the PAN and head network in order to make the model more suitable for mobile devices. It takes advantage of TensorFlow Lite Micro \cite{david2021tensorflow} to quantize the weights and activations with 8-bit precision and finally deploys it on an ultra-low-power microcontroller from the STM32 family. The Authors in \cite{ashfaq2022accelerating} introduced Deeplite Neutrino, which automatically quantizes a CNN model with lower than 4 bits, and presented Deeplite Runtime as an inference engine, which makes the deployment of ultra-low bit quantized models possible on ARM CPUs. Their QAT method can achieve below 4-bit precision for weights and activations of the network, which was rendered feasible with the design of a custom convolution operator using bitserial computation. That is, the calculation of the dot products of the low bit weight and activation values are done through popcount and bitwise operations. They evaluated their method by means of deploying YOLOv5 on Raspberry Pi 4B.
\cite{wang2023infrared} replaces the backbone network with MobileNetV2 \cite{sandler2018mobilenetv2} and adds a coordinate attention mechanism to it. Before deploying the model on NVIDIA Xavier NX, the authors utilized PyTorch to quantize the model after training through a static scheme with 8-bit precision. Similarly, in \cite{cui2022performance}, Pytorch was employed to quantize the modified YOLOv5 in a fake-quantization fashion with 8-bit precision and static clipping range.
The work of \cite{zeng2023lightweight} uses Nihui Convolutional Neural Network (NCNN) framework to quantize the compressed model after training and deploys it on an actual mobile device with a MediaTek Dimensity processor.
\cite{choi2021hardware} proposes a log-scale quantization method that rescales the distribution of activations so that they are suitable for log-scale quantization. This approach minimizes the accuracy drop in YOLOv5 due to log-scale quantization.

Overall, more than half of the reviewed papers used the QAT scheme, which resulted in low precision quantization down to 3 bits. However, none of the PTQ schemes have reached below 8-bit precision. 
While there exists more research on YOLOv5 conducted using quantization, the focus of the review is to primarily include those with novel quantization methods. Therefore, we excluded the results that merely used TensorRT \cite{noauthor_nvidia_2016}, PyTorch Quantization \cite{noauthor_quantization_nodate}, and ONNX quantization \cite{noauthor_quantize_nodate} in their implementation.

\begin{landscape}
\begin{table*}[b]
    \caption{Experimental results of quantization applied on YOLOv5 categorized by deployment scheme. Each scheme method is specified by color. $\Delta$Acc represents the absolute difference in $mAP^{0.5}$ before and after pruning. The rest of the quantitative results represent relative changes. $\Delta$\clock~refers to the changes in the inference time.}
    \centering
    \scriptsize
    \resizebox{\linewidth}{!}{
    \begin{tabular}{C{1.1cm}||C{0.5cm}C{3cm}C{1.2cm}|C{1.2cm}C{1.5cm}C{1.5cm}C{1.5cm}C{1.2cm}|C{0.7cm}C{0.7cm}C{0.8cm}C{1.2cm}}
    \hline
        \textbf{Paper} & \textbf{V} & \textbf{Task} & \textbf{Dataset} & \textbf{Symmetry} & \textbf{Interval} & \textbf{St/Dyn} & \textbf{Fake/IntOnly} & \textbf{Precision} & \textbf{$\Delta$Acc} & \textbf{$\Delta$size} & \textbf{$\Delta$FLOPs} & \textbf{$\Delta$FPS}/\textbf{$\Delta$\clock} \\ \hline
        
        \multicolumn{10}{l}{\textbf{PTQ}} \vspace{.1cm}\\
        \rowcolor{CHANNEL} \cite{zeng2023lightweight} & 5s & Tomato Detection & Manual & --- & --- & --- & --- & FP16 & -0.9 & -51.1 & --- & 18.8/--- \\ 
        \rowcolor{CHANNEL} \cite{zeng2023lightweight} & 5s & Tomato Detection & Manual & --- & --- & --- & --- & Int8 & -5.75 & -73.7 & --- & -5/--- \\
        \rowcolor{CHANNEL} \cite{maji2022yolo} & 5 & Human Pose Estimation & COCO & --- & --- & --- & --- & 8/16 bits & -1.3 & --- & --- & --- \\ 
        \rowcolor{CHANNEL} \cite{wang2023infrared} & 5 & Infrared Object Detection & FLIR ADAS22 & --- & --- & static & --- & Int8 & -1.1 & -75 & --- & 40/--- \\ \hline

        \multicolumn{10}{l}{\textbf{QAT}} \vspace{.1cm}\\
        \rowcolor{HYBRID} \cite{cui2022performance} & 5s & Infrared Ship Detection & Manual & Asym & uniform & static & fake & Int8 & -5 & -27 & -33 & --- \\
        \rowcolor{HYBRID} \cite{wu2023drgs} & 5s & Object Detection & MS-COCO & Asym act/ Sym w & uniform & dynamic & IntOnly & Int4 & -3.1 & --- & --- & --- \\ 
        \rowcolor{HYBRID} \cite{park2022nipq} & 5s & Object Detection & PASCAL VOC & Asym & uniform & dynamic & --- & 5W/5A & -1.2 & --- & --- & --- \\
        \rowcolor{HYBRID} \cite{park2022nipq} & 5s & Object Detection & PASCAL VOC & Asym & uniform & dynamic & --- & 4W/4A & -2.5 & --- & --- & --- \\
        \rowcolor{HYBRID} \cite{park2022nipq} & 5s & Object Detection & PASCAL VOC & Asym & uniform & dynamic & --- & 3W/3A & -5.8 & --- & --- & --- \\
        \rowcolor{HYBRID} \cite{papaioannou2023ultra} & 5 & Crowd Counting & Manual & Asym act/ Sym w & --- & --- & --- & Int8 & -14 & -87.5 & --- & ---/-92.7 \\ 
        \rowcolor{HYBRID} \cite{ashfaq2022accelerating} & 5n & Object Detection & COCO-8class & Asym & uniform & --- & bitserial & $<$Int4 & -1 & --- & --- & ---/-60.6\\ \hline

        \multicolumn{10}{l}{\textbf{Unspecified}} \vspace{.1cm}\\
        \rowcolor{KERNEL} \cite{choi2021hardware} & 5 & --- & --- & Asym & nonuniform & --- & fake & Int8 & -0.9 & --- & -89.2 & --- \\ 
        \rowcolor{KERNEL} \cite{fang2021deployment} & 5 & Object Detection & COCO & --- & --- & --- & --- & FP16 & --- & --- & --- & ---/-60 \\ \hline
        
    \end{tabular}}
    \label{table:QuantizationScheme}
\end{table*}
\end{landscape}

\begin{figure}[t]
    \centering
    \includegraphics[width=0.8\linewidth]{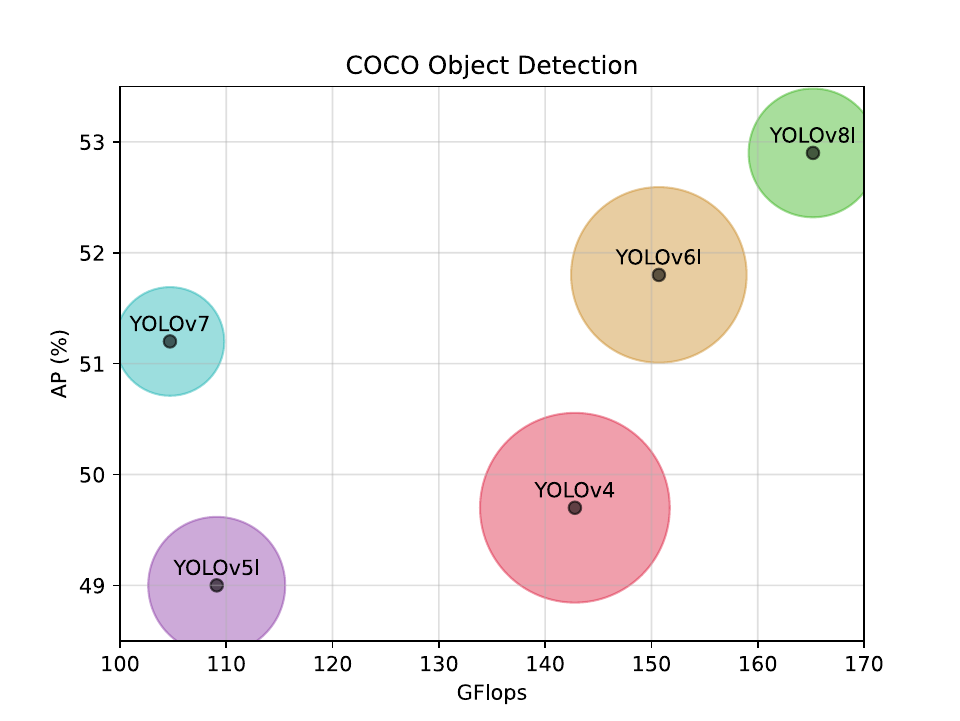}
    \caption{\textbf{Performance characteristic of four different YOLO versions.} The horizontal and vertical axes represent the number of GFlops and the average $AP^{0.5}$ on the COCO 2017 validation dataset with an input size of 640. The radius of circles denotes the relative size of the models.}
    \label{fig:YOLOperformance}
\end{figure}

\section{Conclusions} \label{FutDirection}

Model compression methods have gained significant attention in recent years, and their applications are becoming more specific. In this paper, we presented a review of the commonly used pruning and quantization approaches applied to YOLOv5 and categorized them from different aspects.
Through our analysis, we identified gaps in adapting pruning and quantization methods specifically for YOLOv5. This matter is of importance since the application of those compression methods to YOLOv5 requires further exploration and optimization due to many structural interconnections. Our review serves as a resource for researchers interested in the practical deployment of model compression methods, specifically pruning and quantization, on YOLOv5 and its subsequent newer versions.
With the advent of YOLOv8, YOLO has experienced a new borderline in terms of performance, shown in Figure \ref{fig:YOLOperformance}, which pushes the boundary of object detection and establishes a new SOTA performance.
However, the reduction in terms of size and FLOPs is relatively insignificant which necessitates the need for applying pruning and quantization on newer YOLOs when the limitation in hardware is pivotal. Therefore, our review is extendable to subsequent versions of YOLO and can be used as a guideline for researchers interested in compressing any version of YOLO.

\subsection{Pruning Challenges and Future Directions}
Unlike a regular CNN, pruning YOLOv5 comes with some challenges due to its complex and highly optimized deep neural network architecture.
YOLOv5 uses CSP-Darknet53 neural network architecture as the backbone and PANet as the neck, which both consist of many convolutional layers that are tightly interconnected with concatenations.
Also, the interconnections between the backbone and the neck add to the complexity of the model. Overall, the structural complexity of these layers hinders the process of removing unnecessary filters without adversely affecting the overall structure performance of the network. Otherwise, the spatial resolution of feature maps related to concatenations would not match. Therefore, some compensations should be made prior to pruning the YOLOv5. 
For instance, \cite{zhang2021pruned} does not consider pruning the upsampling layer, the concatenate layer, and the head of the YOLOv5. Also, it ignores the shortcut connection in the BottleNeck module to allow the inputs to have a different number of channels.
In this regard, more studies should consider filter-based and kernel-based pruning because this strategy of pruning does not change the number of output channels; thus, it simplifies the pruning process.\\
As represented in Table \ref{table:pruningGranul}, The current research direction is to utilize the BNSF for sparsity training and channel-based pruning with fine-tuning. However, there is a gap in using one-shot pruning using other saliency criteria. Here we introduce some novel approaches which have not been applied to YOLOv5.\\
EagleEye \cite{li2020eagleeye} treats the pruning process as an optimization problem and states that using the evaluated accuracy might not be a promising criterion to guide the selection of pruning candidates. Therefore, it proposes a random pruning ratio for each layer according to the problem constraints, then prunes the filters based on their $\ell_{1}$-norm. It evaluates the impact of pruning candidates through an adaptive BN-based candidate evaluation module using a sub-sample of training data. In \cite{lin2020hrank}, authors introduced the \textit{HRank filter pruning} which iteratively prunes the filters with low-rank activation maps. \cite{hubara2021accelerated} proposes a \textit{mask diversity} evaluation method to correlate between the structured pruning and the expected accuracy. It also introduces a pruning algorithm called \textit{AdaPrune} that compresses an unstructured sparse model to a fine-grained sparse structured model, which does not need re-training. Similarly, \cite{zhao2019variational} proposes \textit{Variational Bayesian pruning} algorithm that considers the distribution of channels' BN scaling factor instead of deterministically using them similar to section \ref{section:Saliency}.

\subsection{Quantization Challenges and Future Directions}
Although not specific to YOLO, quantizing FP32 to INT8 is not a smooth transformation, and it might hinder the optimality of the results if the gradient landscape is harsh. Also, achieving a low-bit ($<4$~bits) precision using PTQ is almost impossible as it most likely shatters the performance of the model. 
Currently, there is a trend in using ready-to-use quantization modules like TensorRT \cite{noauthor_nvidia_2016}, PyTorch Quantization \cite{noauthor_quantization_nodate}, and ONNX quantization \cite{noauthor_quantize_nodate}, which cannot achieve very low precision as they are limited to 8-bit precision. Nevertheless, such research is not included in this review as our focus is to find novel quantization methods employed on YOLOv5.\\
Regarding applied studies on quantizing YOLOv5, more research is conducted using QAT with a wide range of precision from 1 bit to 8 bits. However, there is a gap in focusing on accelerating training time as well as inference time, especially because training YOLOv5 on a new dataset is computation- and time-consuming. As a solution, more work can be done employing integer-only quantization since hardware throughput is much higher when operations are performed using integer numbers. For instance, TITAN RTX can perform around 23 times more operations per second when the data type is INT4 rather than FP32 \cite{gholami2021sQuanturvey}.
Additionally, PTQ methods still fall back when lower than 8-bit precision is investigated/needed, which presents the opportunity for future work in this area. Hence, we recommend some approaches that can be applied to YOLOv5 in order to fill the above-mentioned gap. In \cite{nagel2020AdaRound}, a PTQ algorithm, AdaRound, is proposed to more efficiently round weights for quantization. It accomplishes SOTA performance with as low as 4-bit precision without a noticeable drop in accuracy ($<1\%$). Yao et al. proposed \textit{HAWQ-V3} \cite{yao2021hawq}, a mixed-precision integer-only quantization method that reaches INT4 or INT4/INT8 uniformly mapped quantization.
\textit{AdaQuant} \cite{hubara2021AdaQuant} proposed a PTQ quantization scheme to minimize the quantization errors of each layer or block separately by optimizing its parameters based on a calibration set. It can attain SOTA quantization with INT4 precision, which leads to a negligible accuracy drop. The authors of \cite{peng2021fully} presented a method for quantization that exclusively utilizes integer-based operations and eliminates redundant instructions during inference. \cite{nahshan2021loss} evaluates the impact of quantization on loss landscape and introduces a novel PTQ approach that can reach 4-bit precision by minimizing the loss function directly, resulting in almost the full-precision baseline accuracy.

\section*{Acknowledgements} We would like to acknowledge the financial support of Sentire and Mathematics of Information Technology and Complex Systems (MITACS) under IT31385 Mitacs Accelerate.


 \bibliographystyle{elsarticle-num} 
 \bibliography{main-ref}





\end{document}